\documentclass{article}

\usepackage{times}
\usepackage{amsmath,amsthm,amssymb}
\usepackage{booktabs}
\usepackage{threeparttable}
\usepackage{epsf}
\usepackage{graphicx}
\newcommand{\be}{\begin{equation}}
\newcommand{\ee}{\end{equation}}
\newcommand{\ba}{\begin{array}}
\newcommand{\ea}{\end{array}}
\newcommand{\bea}{\begin{eqnarray}}
\newcommand{\eea}{\end{eqnarray}}

\renewcommand{\proof}{\noindent {\bf Proof} \quad}
\newcommand{\eproof}{\hfill{$\quad \Box$}}

\newtheorem{thm}{Theorem}[section]

\newtheorem{alg}{Algorithm}[section]

\newtheorem{lem}{Lemma}[section]
\newtheorem{rem}{Remark}[section]

\title{$l_{2,p}-$ Matrix Norm and Its Application in Feature Selection\thanks{The work is partially supported by
the Chinese grants NSFC11001128, NSFC61035003 and NSFC11071117.}}

\author {Liping Wang\thanks{Department of Mathematics,  Nanjing University of
Aeronautics and Astronautics, Nanjing 210016, China.\ Email:
wlpmath@yahoo.com.cn.}\quad\quad Songcan Chen\thanks{Department of Computer Science and Engineering, Nanjing University of Aeronautics and Astronautics, Nanjing, 210016, China.\ Email:
s.chen@nuaa.edu.cn.}}

\begin{document}

\maketitle

\begin{abstract}
Recently, $l_{2,1}$ matrix norm has been widely applied to many areas such as computer vision, pattern recognition, biological study and etc. As an extension of $l_1$ vector norm, the mixed $l_{2,1}$ matrix norm is often used to find jointly sparse solutions. Moreover, an efficient iterative algorithm has been designed to solve $l_{2,1}$-norm involved minimizations. Actually, computational studies have showed that $l_p$-regularization ($0<p<1$) is sparser than $l_1$-regularization, but the extension to matrix norm has been seldom considered. This paper presents a definition of mixed $l_{2,p}$ $(p\in (0, 1])$  matrix pseudo norm which is thought as both generalizations of $l_p$ vector norm to matrix and $l_{2,1}$-norm to nonconvex cases $(0<p<1)$. Fortunately, an efficient unified algorithm is proposed to solve the induced $l_{2,p}$-norm $(p\in (0, 1])$ optimization problems. The convergence can also be uniformly demonstrated for all $p\in (0, 1]$. Typical $p\in (0,1]$ are applied to select features in computational biology and the experimental results show that some choices of $0<p<1$ do improve the sparse pattern of using $p=1$.
\end{abstract}

\section{Introduction}

In many fields, such as computer vision, pattern recognition, computational biology and etc., mixed $l_{2,1}$ matrix norm has received increasing attention for its joint sparsity pattern. In multi-task feature learning, The authors of \cite{1smilarl21} and \cite{2smilarl21} have proposed similar models as $l_{2,1}$-norm regularization to couple feature selection across tasks. But the approach to solve this problem proposed in \cite{3smilarl21} has no known convergence rate. Liu et al. \cite{liujun1} reformulate the nonsmooth $l_{2,1}$-norm regularized optimization to two smooth convex optimization problems, then apply Nesterov's method to solve them. This algorithm analytical computes the solution or globally converges to the solution in linear time. Recently, a proximal alternating direction method is addressed in \cite{he1} to solve $l_{2,1}$-norm regularized least square problem for multi-task feature learning. The $l_{2,1}$-norm involved minimization has also been successfully employed in correlated attribute transfer with multi-task graph-guided fusion \cite{han} and nonnegative graph embedding \cite{zhang1}. Moreover, the authors of \cite{zhao} have used spectral regression with $l_{2,1}$-norm constraint to evaluate features jointly.  The group Lasso \cite{grouplasso,yejieping1} and the logistic group-lasso \cite{logisticlasso} are constructed with $l_{2,1}$-norm regularization in many applications.

One major challenge of $l_{2,1}$-norm minimization is how to efficiently solve this non-smooth optimization problem. The authors of \cite{l21} propose a directly iterative algorithm to solve the robust $l_{2,1}$-norm minimization of both loss function and regularization. And the global convergence is proved in the same literature. The algorithm has been widely used in many applications  for its efficient behavior and construction, for example in \cite{ren,cai1}. This algorithm has been modified to unsupervised feature selection \cite{yy1,li1} and semi-supervised learning \cite{zgma}. A spatial group sparse coding in image-level tagging \cite{yy2} and multi-instance learning \cite{nie2} also employ the similar technique.

On the whole, all the models and algorithms mentioned are constructed in the convex $l_1$-norm framework. Actually, extensive computational studies \cite{lp1,lp2,lp3,xu} have showed that using $l_p$-norm $(0<p<1)$ can find sparser solution than using $l_1$-norm. Naturally, one can expect $l_{2,p}$-norm $(0<p<1)$ based minimization to be a better sparsity pattern than $l_{2,1}$-norm. Recently, a similar $l_p-l_q$ ($0<p\leq 1,\ 1\leq q\leq 2$) penalty for sparse linear and multiple kernel multi-task learning has been considered in \cite{lpq}. But the induced optimization problems have to be separately solved by different algorithms according to the convex ($p=1$) and non-convex ($0<p<1$) cases. This disadvantage brings computational difficulty to freely vary $p$ and $q$. In this paper, we define a mixed  $l_{2,p}$ ($p\in (0, 1]$) matrix norm\footnote{$\|\cdot\|_{2,p}$ ($0<p<1$) is not a valid matrix norm because it does not admit the triangular inequality. Here we call it matrix norm for convenience.} and present a unified algorithm to solve the involved $l_{2,p}$-norm based minimizations for all $p\in (0, 1]$ . To the best of our knowledge, it is the first algorithm to uniformly solve this specially mixed convex and nonconvex optimization problems. The presentation has several innovations as follows. 1) It is a generalization of $l_{2,1}-$norm regularization to nonconvex case. $l_p$-norm ($0<p<1$) is neither convex nor Lipschitz continuous, then the induced $l_{2,p}$-norm based optimization problem is nonconvex and non-Lipschitz continuous yet. 2) Since $l_{2,p}$-norm ($p\in (0, 1]$) based functions are neither convex nor Lipschitz continuous except for $p=1$, efficiently solving the mixed problem is much more challenging than pure $l_{2,1}$-norm minimization. Here we extend the existing work in \cite{l21} to a unified algorithm solving all the $l_{2,p}$-norm $(p\in (0, 1])$ optimization problems. If $p=1$, the general algorithm is reduced to the case of \cite{l21}. If $0<p<1$, the unified algorithm finds a local approximate solution to nonconvex $l_{2,p}$-norm minimization. Fortunately, the convergence can also be uniformly proved for all $p\in (0, 1]$. 3) Typical $p\in (0, 1]$ are tested in $l_{2,p}$-norm based objective functions. The experiments in bioinformatics study provide empirical evidence that some $0<p<1$ are alternatives in constructing sparsity patterns while $p=0.5$ obviously outperforms $p=1$ .

\section{Notations and Definitions}

We employ the notations as usual. Matrices are written as boldface uppercase letters while vectors are written as boldface lowercase letters. For example, $A=(a_{i,j})_{m\times c}$ denotes a real $m\times c$ matrix, $a^i\in R^c (i=1,\cdots, m)$ and $a_j\in R^m (i=1,\cdots, c)$ are the $i-$th row and $j-$th column of $A$ respectively.

For any $x\in R^m$, several useful vector norms are defined as follows,
\be\label{vectornorms}
\|x\|_0=\sum\limits_{x_i\neq 0}|x_i|^0,\quad \|x\|_p^p=\sum\limits_{i=1}^m|x_i|^p,\quad \|x\|_1=\sum\limits_{i=1}^m|x_i|,
\ee
where $p\in (0,1)$. Actually, neither $l_0$ nor $l_p$ ($0<p<1$) is a well defined norm because the former does not satisfy the positive scalability and the latter does not satisfy the triangular inequality. Here we call them norms for simplicity.

$l_{2,1}$-norm of matrix was firstly introduced in \cite{firstl21} which is a strict matrix norm satisfying the norm axioms,
\be\label{matrixnorm1}
\|A\|_{2,1}=\sum\limits_{i=1}^m\|a^i\|_2.
\ee It is well known that $\|\cdot\|_{2,1}$ is convex with respect to matrix variable.
Now we generalize the definition of $l_{2,1}$-norm to mixed $l_{2,p}$-norm as follows
\be\label{matrixnorm2}
\|A\|_{2,p}=(\sum\limits_{i=1}^m\|a^i\|_2^p)^{\frac{1}{p}},\quad p\in (0, 1].
\ee Obviously, $l_{2,p}$-norm is reduced to $l_{2,1}-$norm when $p=1$. Note that $l_{p}$ ($0<p<1$) pseudo norm does not admit the triangular inequality on $R^{m}$, then the corresponding $l_{2,p}$-norm is not a valid matrix norm because of
$$
\|A+B\|_{2,p}\nleq\|A\|_{2,p}+\|B\|_{2,p},\quad A, B\in R^{n\times c}.
$$
Moreover, $l_{p}$ ($0<p<1$) vector norm is neither convex nor Lipschitz continuous, so $l_{2,p}$ matrix pseudo norm is not convex or Lipschitz continuous yet. This properties challenge researchers to uniformly solve the mixed convex and noncovex $l_{2,p}$-norm $(p\in (0, 1])$ based optimization problems.

\section{$l_{2,p}$-Norm Based Minimizations}

Given observation data $\{a_1,a_2,\cdots,a_n\}\in R^d$ and corresponding output $\{b_1,b_2,\cdots,b_n\}\in R^c$, generally principled  framework in many areas is considering
\be\label{eqno0}
\min\limits_{X\in R^{d\times c}}\hbox{loss}(X)+\alpha R(X),
\ee where loss($X$) and $R(X)$ denote loss function and regularization respectively, $\alpha>0$ is the regularization parameter. Different loss($X$) and $R(X)$ are chosen for a variety of data distributions and practical applications. The traditional least square regression solves the following optimization problem to obtain the unknown matrix $X\in R^{d\times c}$:
\be\label{eqno1}
\min\limits_{X}\sum\limits_{i=1}^n\|X^Ta_i-b_i\|_2^2+\alpha R(X),
\ee
where $X$ contains the projection matrix and bias vector for simplicity.

It is well known that the square-norm residual is sensitive to outliers, hence Nie \textit{et. al.} \cite{l21} propose to use robust $l_{2,1}-$norm loss function
\be\label{eqno2}
\min\limits_{X}\sum\limits_{i=1}^n\|X^Ta_i-b_i\|_2+\alpha R(X).
\ee
Here we expect to use the generalized one
\be\label{eqno3}
\min\limits_{X}\sum\limits_{i=1}^n\|X^Ta_i-b_i\|_2^p+\alpha R(X),\quad p\in(0,1].
\ee
For any $p\in (0, 1]$, the noise magnitude of distant outlier in (\ref{eqno3}) is no more than that in (\ref{eqno2}). Thus the model (\ref{eqno3}) is expected to be more robust than (\ref{eqno2}).

Joint sparse regularization of $R(X)$ is usually chosen
\be\label{eqno5}
R_\triangle(X)=\sum\limits_{\|x^i\|_2\neq 0}^d\|x^i\|_2^0\quad \hbox{or}\quad R_\triangledown(X)=\sum\limits_{i=1}^d\|x^i\|_2.
\ee
Theoretically, $R_\triangle(X)$ are mostly preferred for its desirable sparsity. But $R_\triangledown(X)$ is practically chosen more often for the computational sake. Under certain conditions, $R_\triangledown(X)$-regularization is equivalent to $R_\triangle(X)$-regularization. Here we chose the intermediate between $l_0$ and $l_1$ in the sense
\be\label{eqno6}
R_\star(X)=\sum\limits_{i=1}^d\|x^i\|_2^p, \quad p\in(0,1).
\ee
Hence the $l_{2,p}-$norm based feature selection is reduced to a noncovex and non-Lipschitz continuous optimization problem
\be\label{eqno7}
\min\limits_{X}\sum\limits_{i=1}^n\|X^Ta_i-b_i\|_2^p+\gamma^p\sum\limits_{i=1}^d\|x^i\|_2^p,
\ee where $\alpha=\gamma^p$ is the regularization parameter. If $l_{2,1}$-norm based objective are unified in (\ref{eqno7}), it becomes a mixed minimization,
\be\label{eqno77}
\min\limits_{X}\sum\limits_{i=1}^n\|X^Ta_i-b_i\|_2^p+\gamma^p\sum\limits_{i=1}^d\|x^i\|_2^p,\quad p\in (0, 1].
\ee
When $p=1$, problem (\ref{eqno77}) is reduced to the popular $l_{2,1}$-norm based minimization proposed in \cite{l21}. But if $0<p<1$, (\ref{eqno77}) is non-convex, hence the algorithm in \cite{l21} can not be directly applied. As far as we know, very few scheme is presented to uniformly solve this specially mixed problem. Therefore, it is necessary to develop an unified approach to efficiently solve problem (\ref{eqno77}) for all $p\in (0, 1]$.

Denote $A=[a_1,a_2,\cdots,a_n]\in R^{d\times n}$ and $B=[b_1,b_2,\cdots,b_n]^T\in R^{n\times c}$, the objective of problem (\ref{eqno7}) can be written as
\be\label{eqno8}\ba{ll}
J(X):&=\sum\limits_{i=1}^n\|X^Ta_i-b_i\|_2^p+\gamma^p R_\star(X)\\
&=\sum\limits_{i=1}^n\|a_i^TX-b_i^T\|_2^p+\gamma^p\sum\limits_{i=1}^d\|x^i\|_2^p\\
&=\sum\limits_{i=1}^n\|(A^TX-B)^i\|_2^p+\gamma^p\|X\|_{2,p}^p\\
&=\|A^TX-B\|_{2,p}^p+\gamma^p\|X\|_{2,p}^p.
\ea\ee

\section{Main Results}

Obviously, problem (\ref{eqno77}) is equivalent to
\be\label{eqno9}
\min\limits_{X}\frac{1}{\gamma^p}\|A^TX-B\|_{2,p}^p+\|X\|_{2,p}^p.
\ee
Let $E=\frac{1}{\gamma}(A^TX-B)$, then unconstrained optimization problem (\ref{eqno9}) becomes
\be\label{eqno10}\ba{l}
\min\limits_{E,X}\|E\|_{2,p}^p+\|X\|_{2,p}^p,\\
\hbox{s.t.}A^TX-\gamma E=B.
\ea\ee
It can be easily proved that $\|\left[\ba{c}X \\ E\ea\right]\|_{2,p}^p=\|X\|_{2,p}^p+\|E\|_{2,p}^p$. If we denote
\be\label{eqno11}
Y:=\left[\ba{c}X \\ E\ea\right]\in R^{m\times c}\quad\hbox{and}\quad M:=[A^T \ -\gamma I_n]\in R^{n\times m},
\ee where $m=d+n$ and $I_n$ is identity matrix, then problem (\ref{eqno10}) can be reformulated as
\be\label{eqno12}\ba{l}
\min\limits_{Y}\|Y\|_{2,p}^p\\
\hbox{s.t.}MY=B.
\ea\ee

Problem (\ref{eqno12}) is not a convex optimization problem except for $p=1$, so the  solution to (\ref{eqno12}) ($0<p<1$) is a local minimization.  The Lagrangian function of the minimization with linear constraints is
\be\label{eqno13}
\mathcal{L}(Y,\Lambda)=\|Y\|_{2,p}^p-Tr(\Lambda^T(MY-B)).
\ee
where $\Lambda\in R^{n\times c}$ is Lagrangian multiplier matrix, and $Tr(\cdot)$ stands for trace operator.

$Y^\star$ is the KKT point of problem (\ref{eqno12}) if and only if there exists a  $\Lambda^\star\in R^{n\times c}$ such that
\be\left\{\ba{l}\label{eqno14}
\frac{\partial \mathcal{L}(Y,\Lambda)}{\partial Y}=2D_\star Y^\star-M^T\Lambda^\star=0\\
MY^\star=B
\ea\right.,\ee where
\be\label{eqno111}
D_\star=\hbox{diag}\{\frac{p}{2\|y^1\|_{2}^{2-p}},\frac{p}{2\|y^2\|_{2}^{2-p}},\cdots, \frac{p}{2\|y^m\|_{2}^{2-p}}\}
\ee
is induced from $Y^\star$. After simple reformulation, (\ref{eqno14}) is equivalent to
\be\label{eqno15}
Y^\star=D_\star^{-1}A^T(AD_\star^{-1}A^T)^{-1}B.
\ee
If $M$ has full-column rank, then $Y^\star$ satisfying (\ref{eqno15}) is a local minimization to problem (\ref{eqno12}).

Then an iterative algorithm to solve equation (\ref{eqno15}) can be designed as follows.
\begin{alg}\label{alg1}(Solving Problem (\ref{eqno12}))
\begin{enumerate}
\item Start: Given $M\in R^{n\times m}$ and $B\in R^{n\times c}$
\item Set $k=0$ and initialize $D_0=I_m$
\item Iterate: For $k=1,2,\cdots$ until convergence do :
$$
\begin{array}{l}
Y_{k}=D_{k-1}^{-1}M^T(MD^{-1}_{k-1}M^T)^{-1}B,\\
Update\ D_{k}\ with\ diagonal\ entries:\\
\hspace{1cm} \frac{p}{2\|y_{k}^i\|_{2}^{2-p}}, i=1, 2,\cdots, m.
\end{array}
$$
\end{enumerate}
\end{alg}\hfill{$\quad \Box$}

\begin{rem}\label{rem1}
If $D, Y$ are computed as in (\ref{eqno111}) and (\ref{eqno15}), it can be easily derived that Tr$(Y^TDY)=\frac{p}{2}\|Y\|_{2,p}^p$.
\end{rem}
\begin{rem}\label{rem2}
If the $y_{k}^i=0$ happens in some iteration, then $D_k$ can not be well updated and algorithm (\ref{alg1}) breaks down. Here we employ similar techniques in \cite{l21} to overcome it. One choice is setting the $i-$th diagonal element of $D_k^{-1}$ to be $\frac{2\|y_{k}^i\|_{2}^{2-p}}{p}$. Another way is to give a perturbation $\epsilon$ such that $d^{ii}_k=\frac{p}{2\sqrt{(y^i_k)^Ty^i_k+\epsilon}}\neq 0$.
\end{rem}

Now, let us show the convergence of Algorithm (\ref{alg1}). Actually, $\|Y_{k}\|_{2,p}^p$ monotonically decreases with respect to iterations.

\begin{lem}\label{lem1}
If $\varphi(t)=\frac{2}{2-p}t-\frac{p}{2-p}t^{\frac{2}{p}}-1$, where $p\in (0, 1]$, then for any $t>0$, $\varphi(t)\leq 0$.
\end{lem}
\proof Taking derivative of $\varphi(t)$ with respect to $t$, and setting it to zero, that is
$$
\varphi'(t)=\frac{2}{2-p}(1-t^{\frac{2}{p}-1})=0,
$$ then we have the unique stationary point $t=1$ on $(0, +\infty)$.  It can be easily proved that $t=1$ is just the maximum point. Hence
$$
\varphi(t)\leq\varphi(1)=0,\quad t>0.
$$
\eproof
\begin{lem}\label{lem2}
Suppose that $y^i_k$ and $y^i_{k+1}$ are the $i-$th row of $Y_k$ and $Y_{k+1}$ generated by algorithm (\ref{alg1}) respectively, then for $p\in (0, 1]$
\be\label{eqno16}
\|y_{k+1}^i\|_{2}^p-\frac{p}{2}\frac{\|y_{k+1}^i\|_{2}^2}{\|y_{k}^i\|_{2}^{2-p}}\leq \|y_{k}^i\|_{2}^p-\frac{p}{2}\frac{\|y_{k}^i\|_{2}^2}{\|y_{k}^i\|_{2}^{2-p}},\ i=1,\cdots, m.
\ee
Equality in (\ref{eqno16}) holds if and only if $\|y_{k+1}^i\|_{2}^p=\|y_{k}^i\|_{2}^p$.
\end{lem}
\proof
Let $t_\star=\frac{\|y_{k+1}^i\|_{2}^p}{\|y_{k}^i\|_{2}^p}$ in $\varphi(t)$, then $\varphi(\frac{\|y_{k+1}^i\|_{2}^p}{\|y_{k}^i\|_{2}^p})\leq 0$, that is
\be\label{eqno17}
\frac{2}{2-p}\frac{\|y_{k+1}^i\|_{2}^p}{\|y_{k}^i\|_{2}^p}-\frac{p}{2-p}\frac{\|y_{k+1}^i\|_{2}^2}{\|y_{k+1}^i\|_{2}^2}-1\leq 0.
\ee
Note that $\|y_{k+1}^i\|_{2}^p=\|y_{k}^i\|_{2}^p$ is sufficient and necessary to let the equality in (\ref{eqno17}) happen.
Multiplying the two sides of formula (\ref{eqno17}) with $(1-\frac{p}{2})\|y_k^i\|_2^p$, we have
\be\label{eqno18}
\|y_{k+1}^i\|_{2}^p-\frac{p}{2}\frac{\|y_{k+1}^i\|_{2}^2}{\|y_{k}^i\|_{2}^{2-p}}\leq (1-\frac{p}{2})\|y_k^i\|_2^p,
\ee which is also an equivalent formula of (\ref{eqno16}).
\eproof

\begin{thm}\label{thm1}
$\|Y_{k}\|_{2,p}^p$ generated by algorithm (\ref{alg1}) monotonically decreases with respect to iteration $k$. So it converges to the \textsl{KKT} point of problem (\ref{eqno12}) which is also a local minimization of (\ref{eqno12}) if $M$ has full-column rank.
\end{thm}
\proof
From remark (\ref{rem1}) and construction of algorithm (\ref{alg1}), we can easily verify
\be\label{eqno19}
Y_{k+1}=\arg\min\limits_{MY=B}Tr(Y^TD_kY).
\ee
So we have
\be\label{eqno20}
Tr(Y_{k+1}^TD_kY_{k+1})\leq Tr(Y_{k}^TD_kY_{k}),
\ee which is to say
\be\label{eqno21}
\sum\limits_{i=1}^m\frac{p\|y_{k+1}^i\|_{2}^2}{2\|y_{k}^i\|_{2}^{2-p}}
\leq\sum\limits_{i=1}^m\frac{p\|y_{k}^i\|_{2}^2}{2\|y_{k}^i\|_{2}^{2-p}}.
\ee
On the other hand, formula (\ref{eqno16}) in Lemma 4.2 shows
\be\label{eqno22}
\sum\limits_{i=1}^m(\|y_{k+1}^i\|_{2}^p-\frac{p}{2}\frac{\|y_{k+1}^i\|_{2}^2}{\|y_{k}^i\|_{2}^{2-p}})\leq \sum\limits_{i=1}^m(\|y_{k}^i\|_{2}^p-\frac{p}{2}\frac{\|y_{k}^i\|_{2}^2}{\|y_{k}^i\|_{2}^{2-p}})
\ee
Combining equalities (\ref{eqno21}) and (\ref{eqno22}), we have
$$
\sum\limits_{i=1}^m\|y_{k+1}^i\|_{2}^p\leq \sum\limits_{i=1}^m\|y_{k}^i\|_{2}^p,
$$
which is also $\|Y_{k+1}\|_{2,p}^p\leq \|Y_{k}\|_{2,p}^p$. Thus algorithm (\ref{alg1}) generates a monotonically decreasing iterations which converge to the \textsl{KKT} point of problem (\ref{eqno12}). Since  $0<p<1$, problem (\ref{eqno12}) is not a convex optimization. If $M$ has full-column rank, the convergence point of $\{Y_k\}$ is a local minimization of (\ref{eqno12}).
\eproof

\begin{rem}\label{rem3}
To some extent, algorithm \ref{alg1} offers an alternative to solve $l_p$ ($0<p<1$) regularized problems when the number of columns in $Y$ is $1$.
\end{rem}

\begin{rem}\label{rem4}
Algorithm \ref{alg1} is a unified approach to solve problem (\ref{eqno12}) for any $p\in (0, 1]$. This scheme provides algorithmic support to adapt $p$ in $(0, 1]$ to improve sparsity  pattern for different data structure regardless of convex or nonconvex cases.
\end{rem}

It is worth to point out that algorithm \ref{alg1} can be easily extended to solve other general $l_{2,p}$ ($p\in (0, 1]$) regularized minimization
\be\label{eqno23}
\min\limits_{Y\in \mathcal{C}}f(Y)+\sum\limits_{t}\|M_tY+B_t\|_{2,p}^p
\ee by iteratively solving the equivalent form
\be\label{eqno24}
\min\limits_{Y\in \mathcal{C}}f(Y)+\sum\limits_{t}Tr((M_tY+B_t)^TD_k(M_tY+B_t)),
\ee where $D_k=\hbox{diag}\{\frac{p}{2\|(M_tY+B_t)^1\|_{2}^{2-p}},\frac{p}{2\|(M_tY+B_t)^2\|_{2}^{2-p}},\cdots,\\ \frac{p}{2\|(M_tY+B_t)^m\|_{2}^{2-p}}\}$.
Especially consider
\be\label{eqno25}
\min\limits_{Y\in \mathcal{C}}\|A^TY-B\|_F^2+\alpha \|Y\|_{2,p}^p.
\ee The lower bound of nonzero entries in solutions to problem (\ref{eqno25}) is expected to estimate from the theory in \cite{chen}. This possible result is useful to enhance practical algorithm solving problem (\ref{eqno25}).

\section{Experimental Results}

We apply algorithm \ref{alg1} to feature selection in biological study.  In our experiments, four public data sets are used. Brief description about all data sets is given as follows.
\begin{description}
\item[ALLAML] is Leukemia gene microarray data, originally obtained by Golub $et. al.$ \cite{golub10}. There are 7129 genes, containing two classes: acute lymphocytic leukemia (ALL) and acute mylogenous leukemia (AML).
\item[GLIOMA] contains four classes, caner glioblastomas (CG), non-cancer glioblastomas (NG), cancer oligodendrogliomas (CO) and non-cancer oligodendrogliomas (NO). There are total $50$ samples and each class has $14, 4, 7, 15$ samples respectively. Each sample has $12625$ genes.
\item[LUNG] cancer data is available at \cite{gordon}. There are $12533$ genes, total $181$ samples in two classes: malignant pleural mesothelioma (MPM) and adenocarcinoma (ADCA) of the lung.
\item[Prostate-GE] data set has $12600$ genes. There are 102 samples in two classes tumor and normal. $52$ samples are tumor and $50$ samples are normal. The dataset is available in \cite{singh}.
\end{description}
All data set are firstly performed the same preprocessing as in \cite{dudoit}. Then the data sets are standardized to be zero-mean and nomalized by standard deviation. To demonstrate the effect of different $l_{2,p}$ matrix pseudo norms in feature selection, typical $p\in (0, 1]$ are tested by algorithm \ref{alg1}. Here we implement $p=0.25, 0.5, 0.75$ and $1$ in $l_{2,p}$-norm based optimization problems. Using top $20, 40, 60, 80$ features, SVM classifiers are individually performed on all data sets with $5-$fold crosses. The classification errors are reported in  tables \ref{tab1}-\ref{tab2}.

\begin{table*}[!htb]
\centering\small
\caption{\label{tab1}Classification error ($\%$) of different $l_{2,p}$ matrix norms}
 \begin{tabular}{lccccccccc}
 \hline
 & \multicolumn{4}{c}{ Top 20 features}& &\multicolumn{4}{c}{ Top 40 features}\\
\hline
    p=  &  0.25    & 0.5 & 0.75 & 1  & & 0.25    & 0.5 & 0.75 & 1\\ \hline\hline			
  ALLAML & 6.86  &    \textbf{4}      &  6.67 & 5.43 && 5.52  &    \textbf{4.1}         &    5.52 & \textbf{4.1} \\
  GLIOMA & \textbf{0}  &    \textbf{0}        &  \textbf{0}    & 2 && 2  &    \textbf{0}        &  \textbf{0}    & 2\\
  LUNG   & 3.94	&  \textbf{1.98}    & 3.46    & 2.95&& \textbf{1.46} &  \textbf{1.46}   & \textbf{1.46}    & 1.96 \\
  Pro-GE & 4.9 &   \textbf{3.9}     &  6.81   & 5.9 && 8.71		  &   \textbf{6.71}         &  8.71    & 9.71\\
  Average & 3.925 &  \textbf{2.47}  &  4.235 & 4.07 && 4.4225 &  \textbf{3.0675}   &  3.9225    & 4.4425\\
\hline
  \end{tabular}
\end{table*}

\begin{table*}[!htb]
\centering\small
\caption{\label{tab2}Classification error ($\%$) of different $l_{2,p}$ matrix norms}
 \begin{tabular}{lccccccccc}
 \hline
 & \multicolumn{4}{c}{ Top 60 features}& &\multicolumn{4}{c}{ Top 80 features}\\
\hline
    p=  &  0.25    & 0.5 & 0.75 & 1  & & 0.25    & 0.5 & 0.75 & 1\\ \hline\hline			
  ALLAML & 6.86	& \textbf{5.52}   &   6.86 & 8.29 && 8.57	& \textbf{5.71}         &   8.57  & 8.57 \\
  GLIOMA & \textbf{2}	&  \textbf{2}  & \textbf{ 2}    & 4 && 4	&  \textbf{2}         &  \textbf{2 }     & 4 \\
  LUNG   & 9.33	&  \textbf{7.37}   & 8.37    & 10.3&& \textbf{0.99} &  \textbf{0.99}   & 1.48   & 1.48  \\
  Pro-GE & 8.71	&  \textbf{6.71}    &  8.71    & 9.71 && 5.86&  \textbf{3.95 }   &  5.9  & 5.9\\
  Average & 6.725 &  \textbf{5.4}  &  6.485 & 8.075 && 4.855 &  \textbf{3.1625}   &  4.4875  & 4.9875\\
\hline
  \end{tabular}
\end{table*}

The experimental procedure indicates that four $l_{2,p}$-norm ($p=0.25, 0.5, 0.75$ and $1$) based minimizations do select different features, hence result in distinct classification performances. Parameter $p\in(0, 1]$ in $l_{2,p}$ matrix norm balances the sparsity and non-convexity of optimization problem (\ref{eqno12}). The closer to $0$ the $p$ is, the sparser the representation is. While if $p$ is near to $1$, the model is almost convex. The classification error comparisons show that non-convex $l_{2,p}$ $(0<p<1)$ matrix norms provide alternatives to $l_{2,1}$-norm. Especially, $p=0.5$ empirically outperforms $p=1$  in choosing better sparse pattern in various situations.

In order to validate the efficient performance of the unified algorithm \ref{alg1} solving nonconvex $l_{2,p}$ ($0<p<1$) pseudo norm  optimization problems as well as the convex $l_{2,1}$-norm based minimization,  we employ the relative reduction of objective function   $\rho_k=\frac{\|Y_{k}\|_{2,p}^p-\|Y_{k+1}\|_{2,p}^p}{\|Y_k\|_{2,p}^p}$  to estimate the convergence speed. Actually, the convergence behaviors for each $l_{2,p}$-norm case are similar. We display the change of $\rho_k$ with respect to iterative steps in the case of $80$ features (see Figure 1). All experiments on four data sets uniformly get the expected accuracy within around $20$ steps.

\begin{figure}[!htb]\label{fig:convergence}
\centering
\includegraphics[width=55mm]{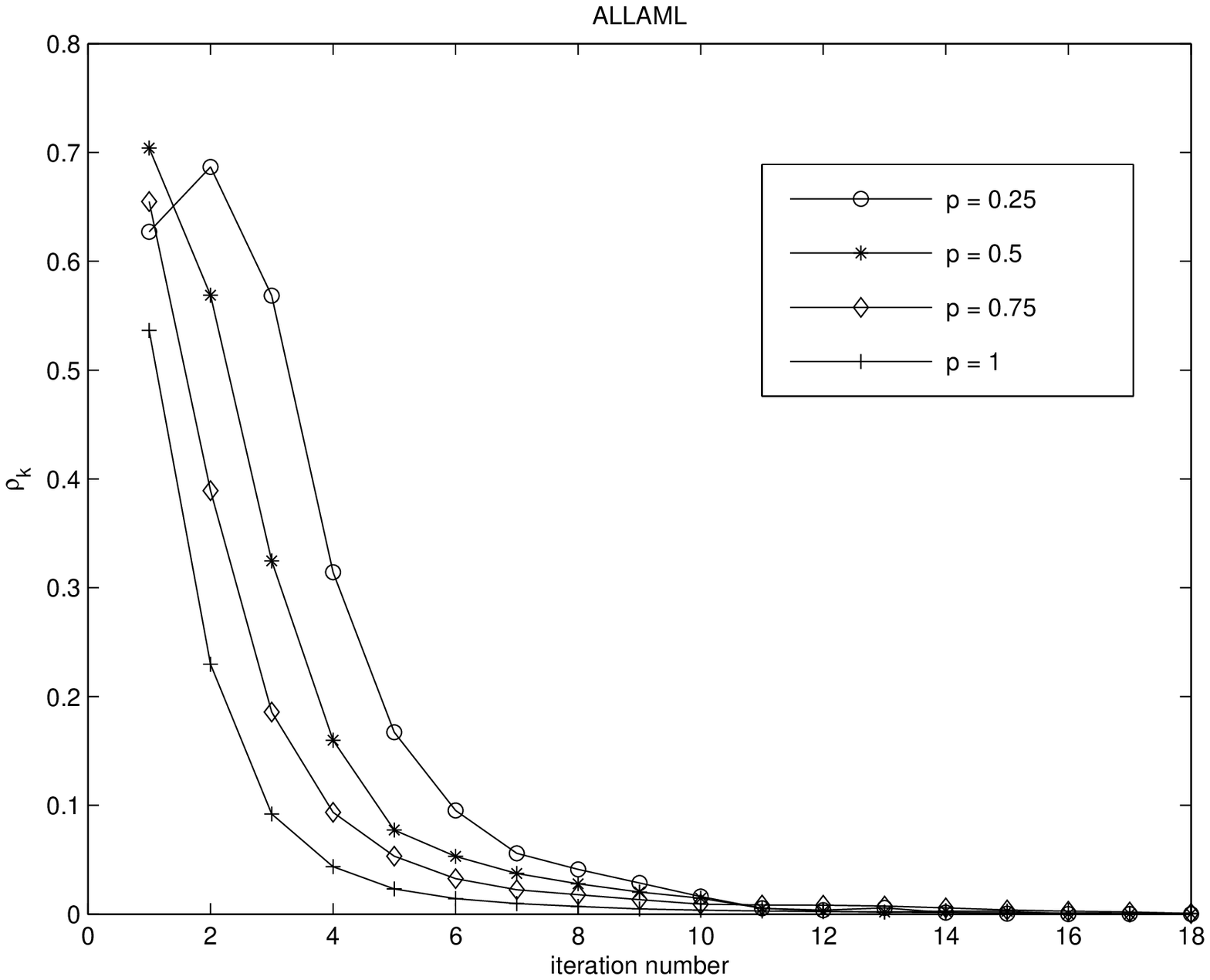}\hspace{3mm}\includegraphics[width=55mm]{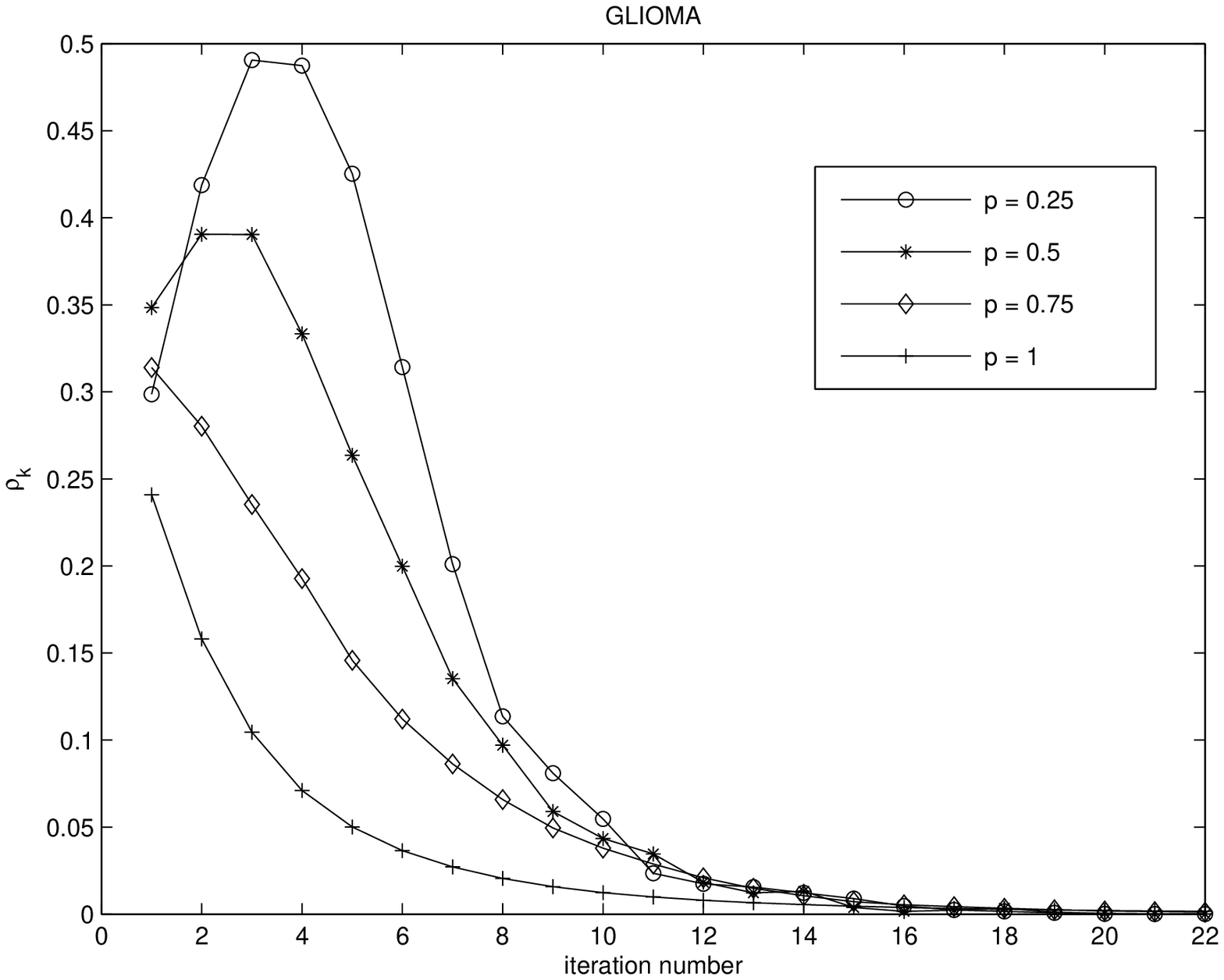}\\
\includegraphics[width=55mm]{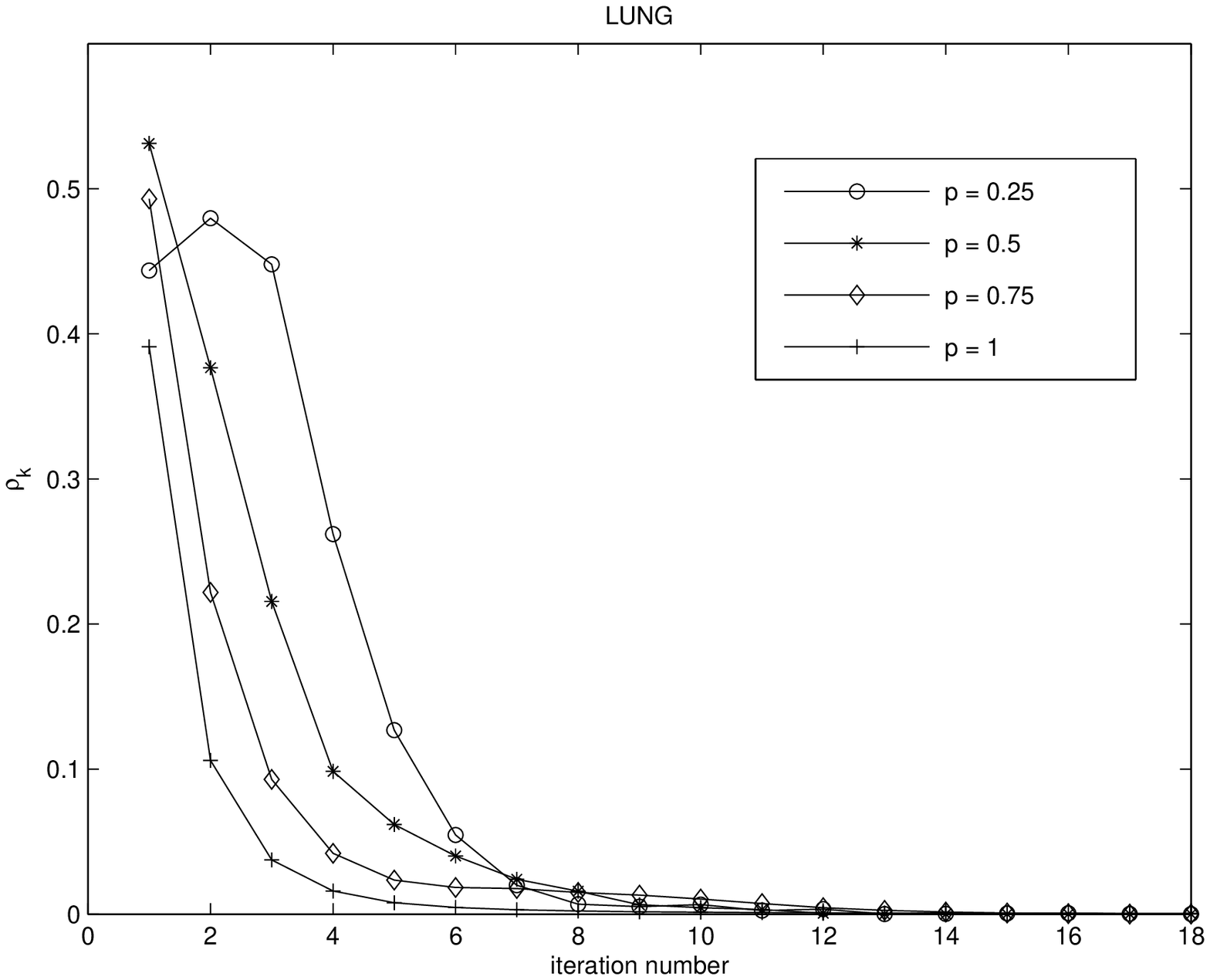}\hspace{3mm}\includegraphics[width=55mm]{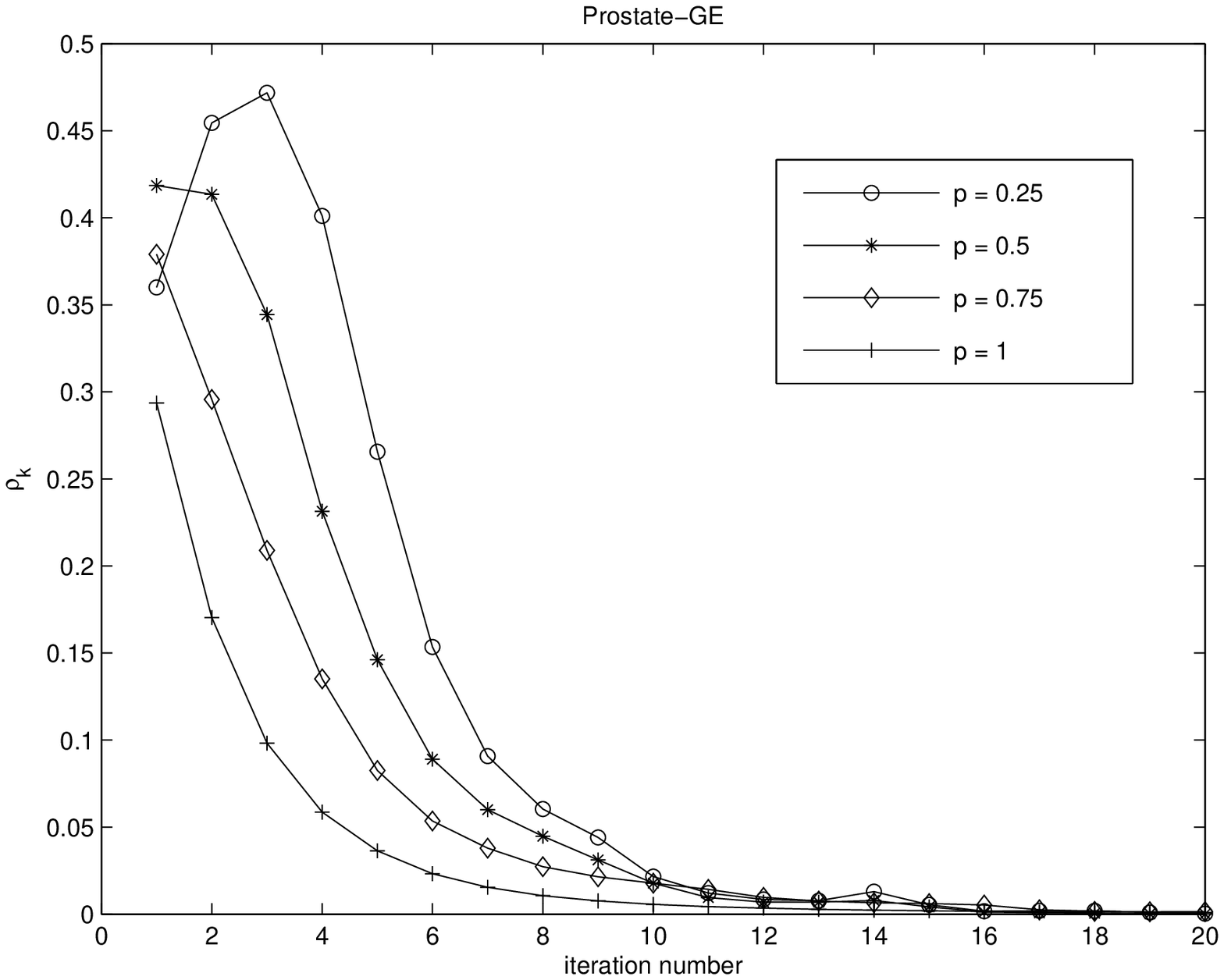}
\caption{The convergence performance of four $l_{2,p}$-norm based minimizations}
\label{fig:equation}
\end{figure}

\section{Conclusions}
In this paper, a kind of general $l_{2,p}$ matrix norms are proposed which are usually used in jointly sparse optimization problems.
A unified algorithm is designed to solve the mixed $l_{2,p}$-norm $(p\in (0, 1])$ based sparse model and the convergence is also uniformly ensured. Experiment results on gene express data sets validate the unified performance of the proposed method. Meanwhile, this approach provides more choices of $p\in (0, 1]$ to  fit variety of jointly sparse structures.


\end{document}